\documentclass[10pt,twocolumn,letterpaper]{article}
 \usepackage{cvpr}              

\usepackage{graphicx}
\usepackage{amsmath}
\usepackage{amssymb}
\usepackage{booktabs}
\usepackage{multirow}
\usepackage{subcaption}
\usepackage{tabularx}
\usepackage{adjustbox}
\usepackage{xcolor}

\usepackage[pagebackref,breaklinks,colorlinks]{hyperref}
\usepackage[capitalize]{cleveref}
\crefname{section}{Sec.}{Secs.}
\Crefname{section}{Section}{Sections}
\Crefname{table}{Table}{Tables}
\crefname{table}{Tab.}{Tabs.}

\newcommand{\datasetname}{ODAH\xspace}
\newcommand{\networkname}{biomechanics-aware network\xspace}
\newcommand{\modelname}{OpenSim skeletal model\xspace}

\newcolumntype{P}[1]{>{\raggedright\arraybackslash}m{#1}}%
\newcolumntype{C}[1]{>{\centering\arraybackslash}m{#1}}%
\newcolumntype{R}[1]{>{\raggedleft\arraybackslash}m{#1}}%
\newcolumntype{Y}{>{\centering\arraybackslash}X}

\definecolor{xucongcolor}{rgb}{0.73725, 0.6588, 0.0705} 
\newif\ifshowcomments
\showcommentstrue 

\usepackage[normalem]{ulem}
\newif\ifshowchange

\usepackage{textcomp}
\usepackage{gensymb}
\usepackage{diagbox}

\begin{document}
\title{3D Kinematics Estimation from Video with a Biomechanical Model and Synthetic Training Data}

 \author{Zhi-Yi Lin\textsuperscript{1}\quad Bofan Lyu\textsuperscript{2} \quad Judith Cueto Fernandez\textsuperscript{2} \quad \\ Eline van der Kruk\textsuperscript{2} \quad Ajay Seth\textsuperscript{2} \quad Xucong Zhang\textsuperscript{1}
 \\ 
     \normalsize\textsuperscript{1}Computer Vision Lab, Department of Intelligent Systems, Delft University of Technology \quad \\
     \normalsize\textsuperscript{2}Department of BioMechanical Engineering, Faculty of Mechanical Engineering (3me), Delft University of Technology\quad \\
 }

\maketitle

\begin{abstract}
Accurate 3D kinematics estimation of human body is crucial in various applications for human health and mobility, such as rehabilitation, injury prevention, and diagnosis, as it helps to understand the biomechanical loading experienced during movement. 
Conventional marker-based motion capture is expensive in terms of financial investment, time, and the expertise required. Moreover, due to the scarcity of datasets with accurate annotations, existing markerless motion capture methods suffer from challenges including unreliable 2D keypoint detection, limited anatomic accuracy, and low generalization capability.
In this work, we propose a novel \networkname that directly outputs 3D kinematics from two input views with consideration of biomechanical prior and spatio-temporal information. To train the model, we create synthetic dataset \datasetname with accurate kinematics annotations generated by aligning the body mesh from the SMPL-X model and a full-body \modelname.
Our extensive experiments demonstrate that the proposed approach, only trained on synthetic data, outperforms previous state-of-the-art methods when evaluated across multiple datasets, revealing a promising direction for enhancing video-based human motion capture.
\end{abstract}
\section{Introduction}
\label{sec:intro}

\begin{figure}[th]
    \centering
    \includegraphics[width=0.9\linewidth]{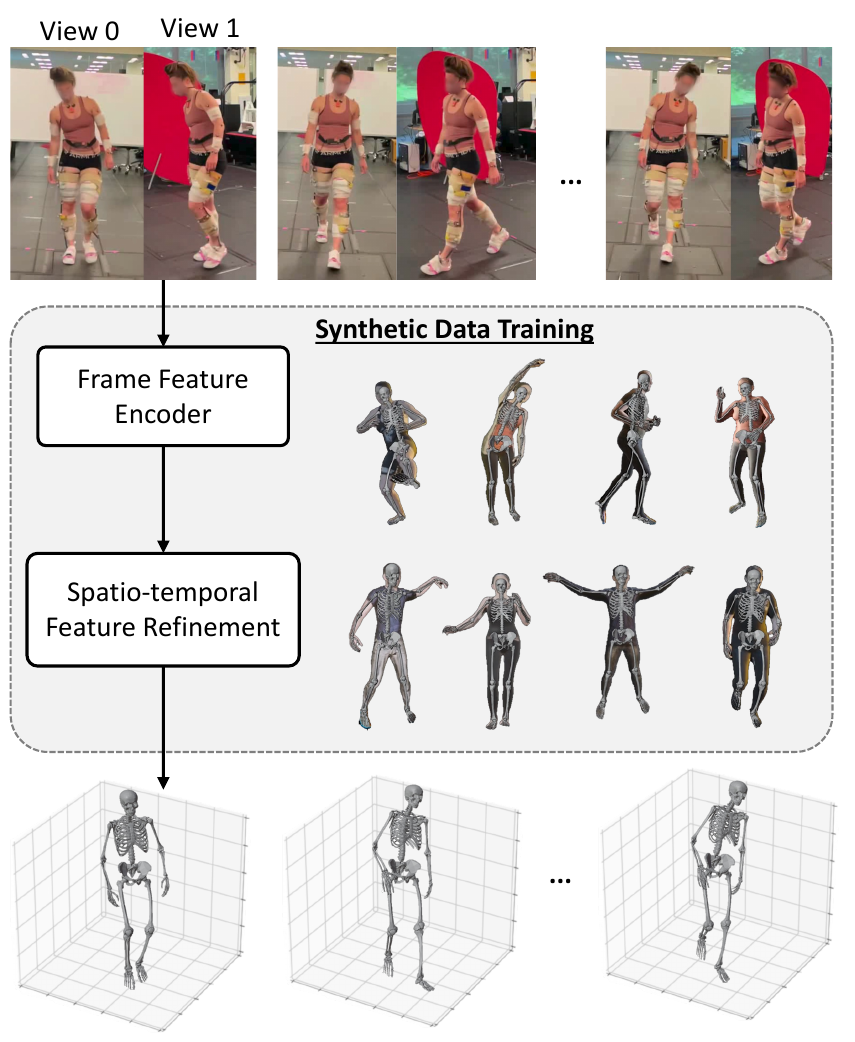}
    \caption{The proposed \networkname consists of a frame feature encoder and a spatio-temporal feature refinement module, which collectively infer 3D kinematics from two-view real-world video inputs. To train the model, we create a synthetic RGB video dataset \datasetname by combining the kinematics skeleton from the OpenSim model, the body mesh from the SMPL-X model, and motions from the AMASS dataset to provide accurate ground truth data.
    Particularly, the end-to-end biomechanics-aware 3D kinematics estimation model is exclusively trained on this self-created synthetic data. Examples of real person images are from the OpenCap dataset \cite{uhlrich2022opencap}, and faces were pixelated for privacy reasons.
    }
    \label{fig:overview}
\end{figure}

Kinematics estimation is the process of capturing the relative position of human body segments through time. In biomechanics, accurate estimation of joint loading of the human body is crucial for applications in healthcare and sports, as it helps in understanding the biomechanical stresses experienced by the joints during movement.
Optoelectronic motion capture systems are the established standard in Kinematics estimation \cite{van2018accuracy}. These systems necessitate the attachment of reflective markers to the bony landmarks of the subject. As a result, this method is time-consuming, requires a substantial financial investment and the markers also restrict the natural movement of the individuals \cite{fleisig2022comparison, mundermann2006evolution}. 
Consequently, many researchers have focused on developing markerless kinematics estimation 
\cite{Pagnon2022JOSS, uhlrich2022opencap, bittner2022towards, song2023markerless}, which usually utilize 3D human pose estimation methods, such as OpenPose \cite{cao2019openpose}, as the backbone to infer 3D poses from detected 2D joints. 
The recent development of deep learning methods has propelled substantial advancements in the field of 3D human pose estimation from detected 2D joints \cite{chen2021anatomy, li2022exploiting, zhang2022mixste, dwivedi2024poco}. It provides great potential 
through the integration of deep learning 3D human pose estimation techniques and biomechanics models. 

Unfortunately, there are significant gaps between the two research fields.
The primary issue is the 2D joint annotations for conventional human pose estimation \cite{cao2019openpose} are often anatomically wrong, resulting in imprecise 3D kinematics estimation. 
Many 3D human pose estimation methods overlook the incorporation of biomechanical constraints, leading to predictions that deviate from anatomically realistic movements. This discrepancy is especially problematic for 3D kinematics estimation, which demands a higher level of biomechanical fidelity for clinical diagnosis and sports science domains.

Another challenge is the scarcity of large-scale datasets with accurate 3D biomechanics annotations. 
The gold standard way to measure human joint kinematics is through the X-ray scan of the person \cite{929617, keller2022osso}, which is not realistic to conduct in daily life settings. 
Although some datasets incorporate maker-based motion capture systems \cite{sigal2010humaneva, uhlrich2022opencap, ghorbani2021movi}, they suffer from sensor noise and lack synchronization with the recording videos. Additionally, the expensive and time-consuming nature of marker-based motion capturing further impedes the construction of large-scale datasets necessary for deep-learning-based markerless motion capture models \cite{bittner2022towards}. 

In this work, we propose a novel markerless motion capture framework comprising a \networkname, as illustrated in \Cref{fig:overview}. The model leverages two-view RGB videos as inputs and performs feature extraction \cite{corona2022learned}, followed by a feature aggregation step to generate frame features. There is no explicit human body joint detection, instead, we sample the feature point on the input frame. Given the dynamic nature of human motion, the frame features are further refined by integrating temporal information through a transformer-based U-Net architecture \cite{zamir2022restormer}. Finally, the model outputs a sequence of joint angles and a set of body segment scales.

To address the limited availability of datasets with accurate kinematics annotations, we propose to create animated humans by aligning the SMPL-X model \cite{SMPL-X:2019} from the computer vision community and OpenSim \cite{delp2007opensim, seth2018opensim} from the biomechanics community. We subsequently animate the SMPL-X model using \modelname and joint angles derived from AMASS dataset \cite{AMASS:ICCV:2019}, which includes diverse motions captured by marker-based motion capture systems. To cover variant situations in real-world settings, we augment the data in terms of clothing, lighting, and camera positions, resulting in a synthetic dataset.
We name this dataset as \textbf{O}penSim \textbf{D}riven \textbf{A}nimated \textbf{H}uman (\datasetname).

Please note we train the proposed \networkname only on our self-created synthetic dataset, which has accurate ground truth labels.
We perform an extensive evaluation of our synthetic data as well as two real-world datasets directly. we demonstrate that our framework outperforms three state-of-the-art markerless motion capture methods on average joint angle error and joint position error across all datasets, which shows the strong generalization of the proposed combination of \networkname and synthetic data.

In summary, the main contributions of this paper are:
\begin{itemize}
\item We introduce an end-to-end 3D kinematics estimation model that predicts joint kinematics and body segment scales with an underlying \modelname.
\item We create a synthetic video dataset \datasetname with accurate kinematics annotations, varied subject appearance, motions, and scene settings.
\item We demonstrate that exclusively trained on synthetic data, that the proposed \networkname can achieve superior performance in average joint angle error across synthetic and real-world datasets, indicating its potential for improving kinematics estimation and domain generalization.
\end{itemize}

\section{Related Works}
\subsection{Markerless Motion Capture with OpenSim}
Markerless motion capture draws people's attention with its cost efficient nature and its ability to yield comparable results to marker-based motion capturing \cite{song2023markerless, mia2023muscles}. Advancements in 3D human motion estimation have enabled the integration of 3D human motion estimation techniques with biomechanical models, allowing for a comprehensive analysis of human biomechanics \cite{keller2023skin}. Most of the existing markerless motion capture methods use multi-step processing for kinematics estimation \cite{Pagnon2022JOSS, uhlrich2022opencap}. The process starts with deriving 3D joint positions by triangulating the detected 2D landmarks \cite{cao2019openpose} from multiple views. Next, the 3D joint positions are treated as marker positions in the Inverse Kinematics (IK) tool and scaling tool in OpenSim software to derive joint kinematics and body segment scales. The mapping from the detected 3D joints to the real 3D joints can be encoded in the marker offsets defined in the OpenSim model \cite{Pagnon2022JOSS}. Alternatively, it is possible to train a model to learn this mapping function \cite{uhlrich2022opencap}. To further improve the performance, the landmark confidence scores are considered to remove low-confidence landmarks \cite{Pagnon2022JOSS, uhlrich2022opencap}.

However, the necessity for operator input and interaction in the intermediate steps (e.g. model scaling and inverse kinematics) of multi-step approaches introduces variability and inconsistency. Therefore, end-to-end solutions are preferred for reliable and accurate 3D kinematics estimation \cite{bittner2022towards, nguyen2023deep}. In addition, the end-to-end method can leverage the raw information contained in the input image and not rely on landmark detection accuracy. D3KE \cite{bittner2022towards}, an end-to-end method, utilizes CNNs to estimate joint kinematics and body segment scales for each frame from monocular videos. A lifting transformer encoder \cite{li2021lifting} is included to refine the predicted joint angles and body segment scales by incorporating temporal information.

Following the end-to-end approach, the proposed model simultaneously estimates joint kinematics and body scales based on visual inputs. Nevertheless, unlike D3KE, the backbone of the proposed network is particularly designed for human pose estimation, two views are utilized to better handle occlusions, and the model is trained on a large-scale synthetic dataset with accurate annotations and varied augmentations.

\subsection{3D Human Pose Estimation}
The two primary approaches are 2D-to-3D lifting and direct 3D estimation. The 2D-to-3D lifting approach requires feature learning from a sequence of 2D poses. This allows for the consideration of temporal information inherent in human motions. Additionally, it compensates for the loss of 3D information from monocular inputs. Temporal dilated Convolution Networks (TCNs) are widely used \cite{pavllo20193d, chen2021anatomy, liu2020attention, xu2020deep} because they can effectively learn spatial and temporal features that are essential for lifting 2D joints and improving motion coherence. In recent years, transformers have gained popularity in handling long-range sequential data \cite{zheng20213d, li2022exploiting, li2022mhformer, zhang2022mixste}. These methods typically involve lifting the 2D pose to 3D ones through transformer-based networks, followed by spatial and temporal refinements.

The direct human motion estimation approach eliminates the reliance on 2D landmark detection by directly predicting 3D poses, offering advantages such as avoiding biases towards specific camera views and mitigating ambiguities associated with 2D landmarks \cite{zhang2021direct, wu2021graph}. Without guidance from 2D landmarks, multi-view inputs, and temporal feature learning become necessary in such frameworks.

With the direct human pose estimation strategy, the proposed method uses a frame feature encoder to map the input frame into the frame features with sampling on the input video frames. Different from previous works \cite{corona2022learned}, we largely improve the architecture to work with two input views to perform 3D kinematic estimation.
In addition, we propose a spatio-temporal refinement module with a transformer-based U-Net architecture adapted from \cite{zamir2022restormer}, which was originally proposed for image denoise.


\section{Method}

\subsection{Overview}
The proposed \networkname first encodes each frame into a frame feature using a stacked hourglass network \cite{newell2016stacked}.
Similar to previous works \cite{uhlrich2022opencap, Pagnon2022JOSS}, we utilize two different views for each frame to provide an absolute 3D kinematic scale, which is essential for many clinical and sports science applications.
Then a spatio-temporal feature refinement is performed to jointly consider spatial and temporal information to optimize the features across frames.
The final output is the joint parameters of the OpenSim model \cite{delp2007opensim}.
To achieve accurate kinematics annotations, we create a synthetic dataset \datasetname with the combination of skeleton from OpenSim \cite{delp2007opensim, seth2018opensim} model, meshes from SMPL-X \cite{SMPL-X:2019}, and acquire the body shape and motion for the mesh from the subject-specific \modelname and joint angles.

\begin{figure}[t]
    \centering
    \includegraphics[width=1\linewidth]{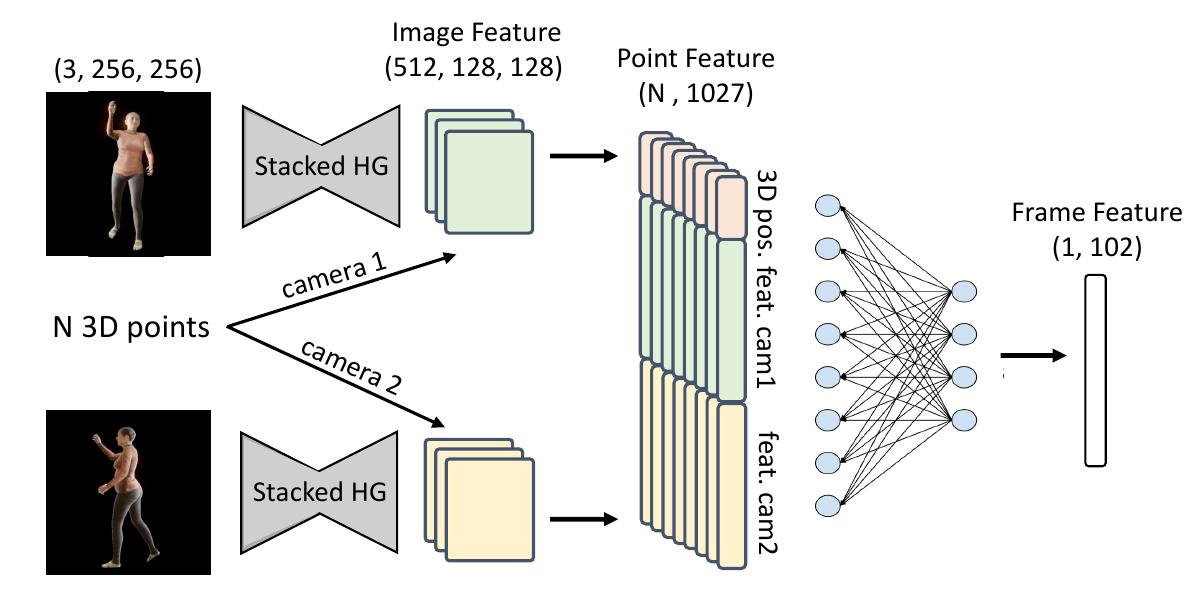}
    \caption{Architecture of the frame feature encoder. Image features are extracted by a stacked hourglass network. The locations to extract the local image features are calculated by projecting the 3D sampled point on two views. Subsequently, point features are derived by concatenating the local image features and the 3D coordinates of the sampled 3D points. Finally, MLP encodes all point features into one compact frame feature.}
    \label{fig:frame_feat_encoder}
\end{figure}

\subsection{Biomechanics-aware network}
\label{subsec:net_arch}
\subsubsection{Frame Feature Encoder}
We first segment the human body using the method proposed in \cite{YOLOv8} and crop the tight bounding box containing the human body.
The \networkname comprises a frame feature encoder and a spatio-temporal refinement module. The frame feature encoder is in charge of generating frame features from two-view frames. More specifically, the image feature of each view in a frame is extracted by a stacked hourglass network, which has demonstrated exceptional performance in human pose estimation tasks \cite{newell2016stacked}. To effectively combine the image features of two views, we first sample candidate 3D points based on the camera rays and the camera parameters. Then, $N$ points are randomly selected from candidates whose 2D projections on both views fall within the human segmentation masks detected by YOLO \cite{YOLOv8}. Last, the point feature, denoted as $\mathbf{z}^{point}_{i} \in \mathbb{R}^{L}$ for point $i$, is obtained by concatenating the 3D point coordinates and the image local features allocated from two views.

To generate one compact frame feature based on the given $N$ point features, we perform a two-stage feature encoding. In the first stage, each point feature is transformed into a more compact representation, denoted as $\tilde{\mathbf{z}}^{point}_{i} \in \mathbb{R}^{L'}$ for point $i$, using a shared MLP. In the second stage, the resulting $N$ compact point features in frame $j$ are concatenated and further encoded into a frame feature $\mathbf{Z}^{frame}_j \in \mathbb{R}^{D}$, for frame $j$, using a second MLP. The architecture of the frame feature encoder is shown in \Cref{fig:frame_feat_encoder}.

\subsubsection{Spatio-temporal Feature Refinement}
\begin{figure}[t]
    \centering
    \includegraphics[width=\linewidth]{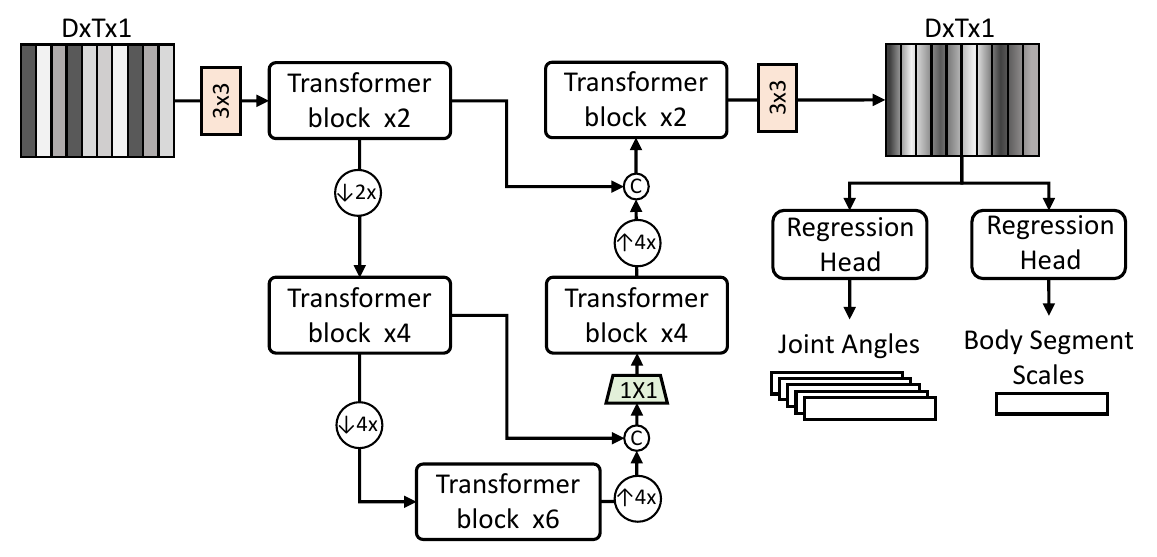}
    \caption{The overview of the proposed spatio-temporal feature refinements. With the sequence of frame features from the frame feature encoder across frames, this refinement architecture treats the feature as a 2D image to process. This process results in a sequence of joint angles and a set of body segments scales with global optimization across frames.}
    \label{fig:sp_tp_refine}
\end{figure}

The spatio-temporal feature refinement is applied to refine a sequence of frame features with temporal information. As shown in \Cref{fig:sp_tp_refine}, this module adopts a transformer-based U-Net architecture to extract multi-range spatio-temporal features from a given sequence of frame features. The adaptations include the downsizing of the U-Net, the temporal-only downsampling and upsampling, and the removal of the skip connection from input to output.

The input to the transformer-based U-Net is generated by concatenating a sequence of frame features along the temporal axis to create a feature map, denoted as $\mathbf{Z}_0^{seq} \in \mathbb{R}^{D \times T \times 1}$, where $T$ is the number of frames in the sequence. The feature map then undergoes processing through the transformer blocks and the U-Net architecture to extract multi-resolution spatio-temporal features along the contracting and expanding paths of the U-Net. During the contracting path, the feature map at level $l$, $\mathbf{Z}_{l}^{seq} \in \mathbb{R}^{D \times t \times c}$, is downsampled to $\mathbf{Z}_{l+1}^{seq} \in \mathbb{R}^{D \times (t/r) \times rc }$, where $r$ represents the downsampling factor from level $k$ to level $l+1$, $t$ is the temporal length, and $c$ is the number of channels. During the expanding path, refined features are generated by hierarchically combining the latent features from different levels using multiple transformer blocks. Finally, regression heads are employed to output a sequence of per-frame joint angles $\boldsymbol{\hat{\theta}} \in \mathbb{R}^{C}$ and the per-sequence body segment scales $\boldsymbol{\hat{s}} \in \mathbb{R}^{B \times 3}$, where $C$ denotes the number of predicted joint angles and $B$ represents the number of body segments.

\subsection{Loss Function}
\label{subsec:loss_func}
For supervision, we utilize a range of factors such as joint angles, body segment scales, biomechanical constraints, and keypoint positions. The final loss function is written as:
\begin{equation} 
\begin{aligned}
\mathcal{L}_{total} = \mathcal{L}_{angle} + \mathcal{L}_{scale} + \mathcal{L}_{bio} + \lambda \cdot \mathcal{L}_{pos}, 
\end{aligned}
\end{equation}
where $\lambda$ is the weight for $\mathcal{L}_{pos}$.

\noindent{\textbf{Joint angles.}}
The joint angles refer to the coordinates defined in the generic \modelname. The angle loss is calculated differently based on whether the rotation of the corresponding joint is constrained. 


The angle loss for free joints, denoted as $\mathcal{L}_{\boldsymbol{\theta}^{f}}$, is calculated as in \cref{eq:angle_dist_free}. This term measures the L1 distance between the predicted and the ground truth when the angles are represented on a unit circle as commonly seen in trigonometry. This representation helps to avoid singularities and angle ambiguities caused by free rotation angles.
\begin{equation} 
\begin{aligned}
\label{eq:angle_dist_free}
\mathcal{L}_{\boldsymbol{\theta}^{f}} & =  \frac{1}{T} \sum_{t=0}^{T-1} \: \lVert \: \hat{\boldsymbol{a}}_{t} - \boldsymbol{a}_{t} \: \rVert_{1},
\end{aligned}
\end{equation}
where $\hat{\boldsymbol{a}}_{t} = (cos\hat{\boldsymbol{\theta}}^{f}_{t}, \: sin\hat{\boldsymbol{\theta}}^{f}_{t}), \: \boldsymbol{a}_{t} = (cos\boldsymbol{\theta}^{f}_{t}, \: sin\boldsymbol{\theta}^{f}_{t})$, $\hat{\boldsymbol{\theta}}^f_t$ and $\boldsymbol{\theta}^f_{t}$ are the predicted and ground truth joint angles of the free joints at time $t$, respectively. $T$ is the number of frames in a sequence.

The angle loss for constrained joints, denoted as $\mathcal{L}_{\boldsymbol{\theta}^c}$, is calculated as the L1 distance between the predicted and ground truth joint angles:
\begin{equation} 
\label{eq:angle_dist_cons}
\mathcal{L}_{\boldsymbol{\theta}^c} =  \frac{1}{T} \sum_{t=0}^{T-1} \: \lVert \: \hat{\boldsymbol{\theta}}^c_{t} - \boldsymbol{\theta}^c_{t} \: \rVert_{1},
\end{equation}
where $\hat{\boldsymbol{\theta}}^c_t$ and $\boldsymbol{\theta}^c_{t}$ are the predicted and ground truth joint angles of the constrained joints at time $t$, respectively. $T$ is the number of frames in a sequence.

\noindent{\textbf{Biomechanical constraints.}}
Predefined constraints are commonly used to ensure biomechanical plausibility by regulating movement. The constraints are further imposed on the network by incorporating $\mathcal{L}_{\boldsymbol{\theta}}^{bio}$ in the loss function to penalize joint angle predictions that violate the constraints as in \cite{spurr2020weakly}. The calculation for $\mathcal{L}_{\boldsymbol{\theta}}^{bio}$ is:
\begin{equation}
\begin{aligned}
\label{eq:angle_range}
\mathcal{L}_{bio} = \frac{1}{T} \sum_{t=0}^{T-1} &  \lVert \: (\hat{\boldsymbol{\theta}}_{t}^{c} \geq \boldsymbol{\theta}_{max}^{c}) \cdot (\hat{\boldsymbol{a}}_{t} - \boldsymbol{a}_{max}) \: \rVert_{1} \\
 & + \lVert \: (\hat{\boldsymbol{\theta}}_{t}^{c} \leq \boldsymbol{\theta}_{min}^{c}) \cdot  (\hat{\boldsymbol{a}}_{t} - \boldsymbol{a}_{min}) \: \rVert_{1},
\end{aligned}
\end{equation}
where $\hat{\boldsymbol{a}}_{t}$, $\boldsymbol{a}_{min}$, and $\boldsymbol{a}_{max}$ are derived as in \cref{eq:angle_dist_free}. $\hat{\boldsymbol{\theta}}_{t}^{c}$ is the predicted joint angles, and [$\boldsymbol{\theta}^c_{min}$, $\boldsymbol{\theta}^c_{max}$] is the allowed range for each joint. The term $T$ denotes the number of frames in a sequence.

\noindent{\textbf{Body segment scales.}}
The body segment scale loss, denoted as $\mathcal{L}_{scale}$, is the L1 distance between predicted and ground truth body segment scales. The calculation is:
\begin{equation} 
\label{eq:scale_loss}
    \mathcal{L}_{scale} = \frac{1}{B} \sum_{i=0}^{B-1} \lVert \hat{\boldsymbol{s}}_{i} - \boldsymbol{s}_{i} \rVert_{1},
\end{equation}
where $\hat{\boldsymbol{s}}_{i}$ and $\boldsymbol{s}_{i}$ represent the predicted and ground truth body segment scales of body segment $i$, respectively. $B$ is the total number of body segments.

\noindent{\textbf{Keypoints.}}
\label{subsub:joint_traj}
We define the position of the joints and mass center of the body segments in the \modelname as the keypoints for loss calculation. Supervision of the keypoint positions implicitly considers the body segment scales and the skeleton topology in 3D space.

Given the joint angles at frame $t$, denoted as $\hat{\boldsymbol{\theta}}_t \in \mathbb{R}^{J}$, and the body segment scales, denoted as $\hat{\boldsymbol{s}} \in \mathbb{R}^{B \times 3}$, the keypoint $\boldsymbol{\hat{P}_t} \in \mathbb{R}^{K \times 3}$ are derived as $\Phi_{forward}(\hat{\boldsymbol{\theta}}_t, \hat{\boldsymbol{s}})$, where $K$ is the number of keypoints, $\Phi_{forward}$ is the kinematics forward function define in the OpenSim model.

The keypoint position loss $\mathcal{L}_{pos}$ is defined as the L1 distance between the prediction and the ground truth keypoint positions. The derivation is: 
\begin{equation} 
\label{eq:pos_loss}
\begin{gathered}
    \mathcal{L}_{pos} = \frac{1}{TK} \sum_{t=0}^{T-1} \sum_{i=0}^{K-1} \lVert \hat{\boldsymbol{p}}_{i, t}  - \boldsymbol{p}_{i, t}  \Vert_{1},
\end{gathered}
\end{equation}
where $\hat{\boldsymbol{p}}_{i, t}$ and $\boldsymbol{p}_{i, t}$ are the predicted and the ground truth keypoint positions, respectively. $T$ is the number of frames, and $K$ is the number of keypoints. Note that all the positions are relative to the pelvis.

\subsection{Synthetic Data Generation}
\label{subsec:syndata}

\begin{figure}[t]
    \centering
    \includegraphics[width=\linewidth]{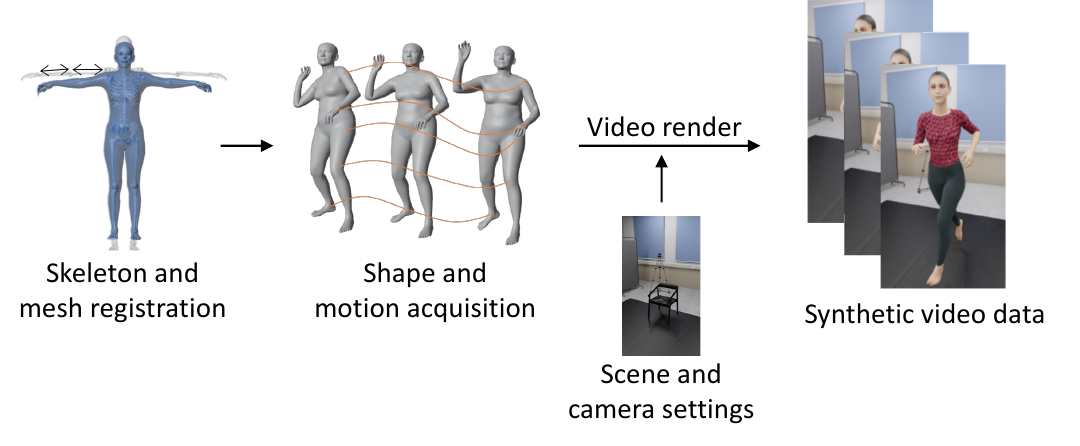}
    \caption{An overview of our synthetic data generation pipeline. We first register the OpenSim skeletal model to the SMPL-X mesh; followed by optimizing the body shape and motion parameters of the mesh to fit the subject-specific \modelname and joint angles. Finally, we simulate real-world environments with scene and camera settings to render the synthetic video data.}
    \label{fig:synthetic_generation}
\end{figure}

We developed a synthetic data generation pipeline to rig an SMPL-X \cite{SMPL-X:2019} model against a full-body \modelname \cite{rajagopal2016full} and its associated joint angles that produce a variety of human motions (Fig \ref{fig:synthetic_generation}). The \modelname and the joint angles serve as the inputs to the pipeline. The subject-specific SMPL-X models were fit to the joint and marker locations extracted from the \modelname. 
In the case of the AMASS dataset, we defined virtual markers on AMASS mesh vertices to obtain marker trajectories. The joint angle sequences were computed from the marker trajectories by scaling and performing inverse kinematics in OpenSim. 

\noindent\textbf{Skeleton and mesh registration.} An OpenSim model is composed of rigid bodies (bones) that are connected by joints. Joints connect two reference frames: one on the parent and one on the child body of the joint, which coincide in space at the joint center. The SMPL-X and \modelname do not share the same joint definitions. 
To remove this discrepancy, we trained a joint regressor for the SMPL-X model that located the joint keypoints corresponding to the OpenSim model joint centers based on the position of the SMPL-X vertices. A variety of poses are included during the training of the joint regressor to avoid overfitting.

\noindent\textbf{Shape and motion acquisition.} We first utilized MoSh++ (Motion and Shape capture) \cite{AMASS:ICCV:2019} to create the initial human mesh. Given marker trajectories and a marker layout on the mesh, MoSh++ generated SMPL-X mesh sequences. The marker layout was defined manually by identifying corresponding SMPL-X vertices as OpenSim markers. From MoSh++, we obtained the initial SMPL-X shape $\beta$ and pose $\theta$ for the given skeletal model and its motion. With additional subject-specific skin or clothing shape for the mesh and a subject-specific OpenSim skeleton, we further optimized $\beta$ and pose $\theta$ to minimize differences between virtual markers on the mesh and the markers on the skeletal model. 


To animate the resulting subject-specific SMPL-X mesh, its pose parameters, $\theta$ are optimized frame by frame to minimize the distance between joint keypoints and marker positions of the SMPL-X mesh and those on the \modelname driven by the input joint angles.



\noindent\textbf{Scene settings.} To augment appearances, we used four types of upper body clothing, ranging from vests to long-sleeved shirts, and four types of lower body clothing, from shorts to trousers. Each type of the clothing is randomly combined with 5 different textures in each trial. We used multiple area light sources evenly distributed on the ceiling.

\noindent\textbf{Cameras settings.} We employed two static cameras for video rendering that were positioned at a height of 1.1 $\pm$ 0.1 meters. One camera captured the frontal view, while the other one captured the sagittal view. To enhance diversity, the positions of the cameras are randomly perturbed within a small range. Both cameras had a fixed focal length of 33 mm. The sensor fit of the cameras was set to horizontal with the sensor width set to 36 mm. 

\noindent\textbf{Rendering.} The videos were rendered using the Metal-accelerated BLENDER\_EEVEE engine in Blender 3.5. The video resolution was set to 1080 by 720 and collected at a framerate of 60 fps. Motion blur effects are disabled. The videos are encoded in the H264 format with a medium-quality configuration.


\section{Experiments}

\subsection{Datasets}
\noindent{\textbf{\datasetname.}}
In the proposed \datasetname, there are 56 synthetic subjects generated from the BMLMovi \cite{ghorbani2021movi} in a subset of the AMASS \cite{AMASS:ICCV:2019} dataset. The actions include running, jogging, jumping, sideways, scratching head, throwing and catching, hand clapping, walking, checking watch, sitting down, hand waving, crossing arms, stretching, kicking, phone talking, taking photos, pointing, vertical jumping, crawling, crossing legs while sitting, and freestyle. Overall, \datasetname has 1132 videos in 60 fps, and each video has a duration of around 10 seconds. We use 42 subjects for training, six subjects for validation, and eight subjects for testing.

\begin{table*}[t]
\centering
\begin{tabularx}{0.9\textwidth}{l|YYYY|YYYY}
\toprule
  & \multicolumn{4}{c}{$MAE_{angle}$ \text{(deg.)} $\downarrow$}                     \vline     & \multicolumn{4}{c}{PA-MPJPE \text{(mm)} $\downarrow$} \\
 & Pose2Sim       & OpenCap       & D3KE  & Ours           & Pose2Sim       & OpenCap        & D3KE  & Ours  \\ \hline
OpenCap  & 9.77    & \textbf{7.37} & 12.47 & 9.65          & 67.69          & \textbf{53.86} & 95.64          & 74.85         \\
BMLmovi  & 10.68    & 15.27        & \textbf{9.43} & 9.73 & 58.55          & 113.45        & \textbf{49.31} & 64.68         \\
\datasetname  & 10.99    & 10.78       & 12.11 & \textbf{4.81}  & 60.67         & 64.27         & 74.29         & \textbf{27.09} \\ \hline
Mean     & 10.48    & 11.14         & 11.34 & \textbf{8.06}  & 62.30 &  77.19       & 73.08        & \textbf{55.54} \\
\bottomrule
\end{tabularx}
\caption{Comparison between Pose2Sim, OpenCap, D3KE, and our method (columns) with the joint angle error (MAE) and joint keypoint error (PA-MPJPE). The evaluation is performed on OpenCap, BMLMovi, and \datasetname (rows) datasets. The last row shows the average error across all datasets.}
\label{tab:result_all}
\end{table*}

\noindent{\textbf{Testing.}}
To evaluate the performance of the proposed \networkname and the generalization capability in real-world settings, we tested the method on two real datasets, OpenCap \cite{uhlrich2022opencap} and BMLMovi \cite{ghorbani2021movi}, as well as the proposed synthetic \datasetname.
OpenCap consists of ten subjects performing actions including walking, squatting, rising from a chair, drop jumps, and the asymmetric counterparts. It was recorded using five RGB cameras and a marker-based motion capture system. OpenCap also provides processed marker data and kinematics annotations for a full-body \modelname. BMLMovi involves 90 subjects performing 21 actions, recorded using two cameras and a marker-based motion capture system. Since BMLmovi does not include kinematics annotations, we utilized the provided marker data, the OpenSim Scale tool, and the OpenSim IK tool to acquire accurate joint angles and body segment scales for ground truth. We recognize the root-mean-square of the marker data and the fitted ground truth is between one to three centimeters. We excluded two actions, namely crawling and crossing legs while sitting, from our test set due to the challenges in fitting the underlying \modelname to the bending movement.

\subsection{Implementation Details}
We first train the stacked hourglass network for 14 epochs with a batch size set to eight, and then integrate it with the spatio-temporal refinement network for another five epochs with a batch size set to two. 
We use the Adam optimizer \cite{kingma2014adam} with $\beta_1 = 0.5$ and $\beta_2 = 0.999$ is applied, and the learning rate is set to $1\times5 \times 10^{-5}$. Input frames are resized to $3\times256\times256$ before feature extraction. The loss weight $\lambda$ is set to 100. We train our method with a single A40 GPU.

We set $N=500$ 3D points for point feature extraction, the point feature length $L=1027$, and the reduced one $L'=32$. The final frame feature length $D=102$. The sequence length $T$ is set to $64$. The U-Net consists of three encoder-decoder levels, with downsampling factors $r$ set to two for level two and four for level three. The number of transformer blocks for each level is [2, 4, 6], the number of attention heads is [1, 2, 4], and the number of channels is [48, 96, 384].

In the generic \modelname, the number of the joint angles $J$ is $36$. The number of body segments $B$ is 22, and the number of joint keypoints $K$ is 44. Our generic \modelname has nine unconstrained joint angles, controlling the pelvis, and left and right arms. Therefore, only 17 joint angles are restricted by biomechanical constraints.

\subsection{Metrics}
To make the evaluation focus more on the performance of the 3D joint kinematics, Procrustes alignment \cite{gower1975generalized} is applied as the first step to align the global translation, rotation, and scaling between the predictions and the ground truth.

Mean Absolute Error (MAE) is used to evaluate joint angle error as 
\begin{equation} 
\label{eq:joint_err}
\begin{gathered}
   MAE_{angle} = \frac{1}{T} \sum_{t=1}^{T} \lVert \hat{\boldsymbol{\theta}}_{t} - {\boldsymbol{\theta}}_{t} \rVert_{1},
\end{gathered}
\end{equation}
where $\hat{\boldsymbol{\theta}}_{t}$ is the predicted angles, $\boldsymbol{\theta}_{t}$ is the ground truth angles, and $T$ is the number of frames in a sequence.

Mean Per Joint Position Error with Procrustes Alignment (PA-MPJPE) is widely used in 3D human motion estimation to measure the Euclidean distance between the predicted and the ground truth 3D joint keypoint positions. The calculation is
\begin{equation}
\label{eq:3dhpe_mpjpe}
PA{\text -}MPJPE = \frac{1}{TB} \sum_{t=0}^{T-1} \sum_{i=0}^{B-1} \lVert  \hat{\boldsymbol{p}}_{i, t}^{PA}  - \boldsymbol{p}_{i, t} \Vert_{2}, 
\end{equation}
where $B$ is the number of joint keypoints, $\hat{\boldsymbol{p}}_{i, t}^{PA}$ is the predicted joint keypoint positions relative to the root joint after Procrustes alignment and $\boldsymbol{p}_{i, t}$ represents the ground truth joint keypoint positions relative to the root joint.

\subsection{Baselines}

We compare the performance of the proposed method against state-of-the-art methods of two multi-step methods Pose2Sim \cite{Pagnon2022JOSS} and OpenCap \cite{uhlrich2022opencap}, and one end-to-end method D3KE \cite{bittner2022towards}. 

For Pose2Sim and OpenCap, frontal and sagittal views are taken as inputs and the Body25 model in OpenPose \cite{cao2019openpose} is used as the 2D landmark detection backbone. We use Pose2Sim default Butterworth low-pass filter with a cut-off frequency of six Hz as the smoothing filter. For OpenCap, the cut-off frequency is set to half of the framerate. To ensure a fair comparison, we disable the landmark synchronization and the video trimming in OpenCap. For D3KE, it only requires a single-view image as input that we take frontal view. We acknowledge that it is not a totally fair comparison due to different inputs. However, it is natural of the D3KE method with only single-view as the input. We choose the transformer-based temporal model with the sequence length set to 243 frames for the D3KE method. Additionally, we only compare joint angles present in our generic \modelname and exclude arm flexion since D3KE's generic \modelname does not have a joint angle defined for arm flexion for the sake of fair comparison.

\subsection{Comparison with State-of-the-Art}
We compare the proposed \networkname with state-of-the-art methods and show the results in \Cref{tab:result_all}. We conduct experiments on three datasets OpenCap, BMLmovi, and our synthetic \datasetname as shown in each row, and compare three methods Pose2Sim, OpenCap, and D3KE as shown in columns. We report the errors on both joint angle error ($MAE_{angle}$) and joint keypoint position error (PA-MPJPE). We also average the performance across three datasets for each method for comparison. 
From the table, we can see that our method achieves the best performances in terms of averaged joint angle and keypoint position errors. Our method is significantly better than the second-best baseline Pose2Sim with 23\% improvement (8.06 degrees vs. 10.48 degrees) on joint angle estimation error and 10\% improvement (55.54 mm vs. 62.30 mm) on the joint keypoint position error.
It clearly shows the advantage of the proposed model over the previous state-of-the-art methods with our advanced neural network architecture and spatial-temporal feature refinement.

Note that the OpenCap method was trained on the OpenCap training dataset, D3KE was trained on the BMLmovi dataset training set, and \networkname was trained on \datasetname training set. Therefore, each of these methods achieves the best performance on the corresponding test set with their specific domain knowledge.
However, our \networkname shows strong generalizability by achieving the best averaged performances across the three datasets.
Even if we remove results on our synthetic dataset \datasetname from the \Cref{tab:result_all}, our \networkname still can achieve
the best averaged performances across the rest two real-world datasets in terms of the joint angle estimation task.
Since \networkname is solely trained on synthetic \datasetname without finetuning on any real data, the superior average results show the effectiveness of using synthetic data 
improving domain generalization. These results also confirm the biomechanical fidelity of the proposed synthetic dataset \datasetname. 

\subsection{Ablation Study}
\begin{table}[t]
\centering
\begin{adjustbox}{width=\columnwidth,center}
\begin{tabular}{l|ccc|ccc}
\toprule
    & \multicolumn{3}{c}{$MAE_{angle}$ \text{(deg.)} $\downarrow$}  \vline    & \multicolumn{3}{c}{PA-MPJPE \text{(mm)} $\downarrow$}  \\
  & small    & medium    & large    & small    & medium    & large \\ \hline
OpenCap   & 11.04 & 10.64  & \textbf{9.65} & 92.62 & 81.79          & \textbf{74.85} \\
BMLmovi   & 10.54 & 9.95  & \textbf{9.73} & 82.91 & 69.53 & \textbf{64.68}          \\
\datasetname   & 6.23  & 5.39   & \textbf{4.81}  & 39.71 & 33.60         & \textbf{27.09} \\ \hline
Mean      & 9.27 & 8.66   & \textbf{8.06}  & 71.75 & 61.64          & \textbf{55.54}   \\
\bottomrule
\end{tabular}
\end{adjustbox}
\caption{Ablation study on different data sizes. We configure \datasetname into small, medium, and large subsets. The evaluation is performed on OpenCap, BMLMovi, and \datasetname (rows) with joint angle error (MAE)
and joint position error (PA-MPJPE).}
\label{tab:abl_datasize}
\end{table}

\begin{table}[t]
\begin{adjustbox}{width=\columnwidth,center}
\centering
\begin{tabular}{l|cc|cc|cc}
\toprule
 & \multicolumn{2}{c}{$MAE_{angle}$ \text{(deg.)} $\downarrow$}                     \vline     & \multicolumn{2}{c}{PA-MPJPE \text{(mm)} $\downarrow$}  \vline & \multicolumn{2}{c}{PA-MPJVE \text{(mm/s)} $\downarrow$} \\
 & Frame & Ours & Frame & Ours  & Frame & Ours  \\ \hline
OpenCap  & 9.91 & \textbf{9.65} & 76.59 & \textbf{74.85} & 906.65 & \textbf{471.37} \\
BMLmovi  & 9.82 & \textbf{9.73} & 65.59 & \textbf{64.68} & 394.71  & \textbf{254.02} \\
\datasetname  & \textbf{4.61}  & 4.81          & 27.36 & \textbf{27.09} & 296.45  & \textbf{155.71} \\ \hline
Mean     & 8.11  & \textbf{8.06}  & 56.51 & \textbf{55.54} & 532.60  & \textbf{293.70}\\
\bottomrule
\end{tabular}
\end{adjustbox}
\caption{Ablation study on the effectiveness of temporal information. Frame-based prediction is the proposed network without spatio-temporal refinement, and we compare this baseline to our final model. The evaluation is performed on OpenCap, BMLMovi, and \datasetname (rows) with joint angle error (MAE), joint position error (PA-MPJPE), and joint velocity error (MPJVE).}
\label{tab:abl_frameseq}
\end{table}

\begin{table}[ht]
\centering
\begin{adjustbox}{width=\columnwidth,center}
\begin{tabular}{l|ccc|ccc} 
\toprule
       & \multicolumn{3}{c}{$MAE_{angle}$ \text{(deg.)} $\downarrow$}                      \vline    & \multicolumn{3}{c}{PA-MPJPE \text{(mm)} $\downarrow$}  \\ \hline

$\mathcal{L}_{angle}$ & \checkmark  & \checkmark  & \checkmark  & \checkmark  & \checkmark  & \checkmark    \\
$\mathcal{L}_{scale}$  & \checkmark  & \checkmark  & \checkmark  & \checkmark  & \checkmark  & \checkmark   \\
$\mathcal{L}_{bio}$    &       & \checkmark    & \checkmark     &       & \checkmark    &  \checkmark     \\
$\mathcal{L}_{pos}$    &       &      & \checkmark   &       &      & \checkmark  \\\hline
OpenCap              & 10.68         & 10.20 & \textbf{9.65}  & 90.74 & 92.64 & \textbf{74.85}        \\
BMLmovi              & 9.78         & 9.75 & \textbf{9.73} & 78.51 & 84.09 & \textbf{64.68}          \\
\datasetname              & \textbf{4.45} & 4.68  & 4.81             & 36.45 & 37.24  & \textbf{27.09}          \\ \hline
Mean                 & 8.30 & 8.21  & \textbf{8.06}            & 68.57 & 71.32 & \textbf{55.54}   \\ 
\bottomrule
\end{tabular}
\end{adjustbox}
\caption{Ablation study on the loss function. We incrementally test the effects of biomechanical constraints (+ $\mathcal{L}_{angle}$) and the keypoint positions (+ $\mathcal{L}_{pos}$). The evaluation is performed on OpenCap, BMLMovi, and \datasetname (rows) with joint angle error (MAE) and joint position error (PA-MPJPE).}
\label{tab:abl_loss}
\end{table}

\subsubsection{Training Data Size}
The benefit of synthetic data generation is its great potential for large amounts of data creation. To verify the effectiveness of the proposed \datasetname, we configure the \datasetname into three subsets. The small set has 10 subjects and 195 clips, the medium set has 20 subjects and 397 clips, and the large set has 42 subjects and 841 clips. Note the large set is the full \datasetname dataset. We train our model on each of these subsets and test on the test set of \datasetname. 
We show results in \Cref{tab:abl_datasize} across three datasets as each row and training subsets as each column.
From the table we can that the performances of the large subsets achieve the best performances across all datasets, thus, indicating that more data ensures better performances for kinematics estimation in terms of both joint angles and position. Note the large subset training does not cause the overfitting problem on the two real-world datasets OpenCap and BMLmovie. It confirms again the high biomechanical fidelity of the proposed \datasetname which benefits the model training for the generalizability. Since it is feasible to generate large synthetic data with the proposed synthetic data generation method, there is great potential for future method development.

\subsubsection{Frame-based vs. Sequence-based}
In the method design, we have two components including the frame feature encoder for the frame-based prediction and the spatio-temporal feature refinement for the sequence-based prediction.
Essentially, the frame-based prediction can already perform the 3D kinematic estimation task.
To investigate the effectiveness of feature refinement, we perform a comparison between the frame feature encoder only and the completed model. Specifically, frame-based prediction is implemented by removing the spatio-temporal refinement module in the proposed framework, and directly outputs the joint angles and body segment scales. 
For experiments, besides the joint angle and position errors, we also calculate the Mean Per Joint Velocity Error (MPJVE) since it indicates the smoothness of the motion.
The results are shown in \Cref{tab:abl_frameseq}, which indicates that the completed model achieves better results than the frame-based prediction baseline in general.
Moreover, the improvements in the motion coherence, as indicated by Mean Per Joint Velocity Error (MPJVE), are much more significant.

\subsubsection{Loss Function}
Finally, We further conduct tests to examine the contribution of each loss term.
The joint angle loss $\mathcal{L}_{angle}$ and body segment scale loss $\mathcal{L}_{scale}$ are essential for the training that must be included. Therefore, we only examined the effectiveness of biomechanical constraints $\mathcal{L}_{bio}$ and keypoint position loss $\mathcal{L}_{pos}$.
The results are present in \Cref{tab:abl_loss}. 
Although the $\mathcal{L}_{bio}$ does not bring significant improvement numerically, we notice there could be extreme joint angle and position predictions without it. The implicit weights on joint angles introduced by $\mathcal{L}_{pos}$ can improve both the joint angle error and the joint keypoint position error. Combining all loss terms, our model can achieve the optimal performance across the three datasets. 

\section{Conclusion}
We propose a novel end-to-end \networkname that is solely trained on self-created synthetic data.
The proposed method utilizes direct mapping from two input views to the frame features and is refined with a spatio-temporal refinement module.
We create synthetic data by combining the SMPL-X model from the computer vision community and the \modelname from the biomechanics community to provide accurate ground truth of kinematics.
By solely trained on synthetic data, our proposed method achieves superior performance with the best generalization across multiple datasets.
It demonstrates the effectiveness of the proposed method for improving kinematics estimation and enhancing domain generalization, and also validates the biomechanical fidelity of the proposed dataset generation pipeline.

The limitations include visual quality and motion variations in the synthetic dataset, and the size of the proposed architecture. Future research could focus on improving the visual quality via adversarial training, including variant actions from real humans, and extending the method.

{\small
\bibliographystyle{ieee_fullname}
\bibliography{08_reference.bib}

\begin{thebibliography}{10}\itemsep=-1pt

\bibitem{bittner2022towards}
Marian Bittner, Wei-Tse Yang, Xucong Zhang, Ajay Seth, Jan van Gemert, and Frans~CT van~der Helm.
\newblock Towards single camera human 3d-kinematics.
\newblock {\em Sensors}, 23(1):341, 2022.

\bibitem{cao2019openpose}
Zhe Cao, Gines Hidalgo, Tomas Simon, Shih-En Wei, and Yaser Sheikh.
\newblock Openpose: Realtime multi-person 2d pose estimation using part affinity fields, 2019.

\bibitem{chen2021anatomy}
Tianlang Chen, Chen Fang, Xiaohui Shen, Yiheng Zhu, Zhili Chen, and Jiebo Luo.
\newblock Anatomy-aware 3d human pose estimation with bone-based pose decomposition.
\newblock {\em IEEE Transactions on Circuits and Systems for Video Technology}, 32(1):198--209, 2021.

\bibitem{mia2023muscles}
Mia Chiquier and Carl Vondrick.
\newblock Muscles in action.
\newblock In {\em ICCV}, 2023.

\bibitem{corona2022learned}
Enric Corona, Gerard Pons-Moll, Guillem Aleny{\`a}, and Francesc Moreno-Noguer.
\newblock Learned vertex descent: a new direction for 3d human model fitting.
\newblock In {\em Computer Vision--ECCV 2022: 17th European Conference, Tel Aviv, Israel, October 23--27, 2022, Proceedings, Part II}, pages 146--165. Springer, 2022.

\bibitem{delp2007opensim}
Scott~L Delp, Frank~C Anderson, Allison~S Arnold, Peter Loan, Ayman Habib, Chand~T John, Eran Guendelman, and Darryl~G Thelen.
\newblock Opensim: open-source software to create and analyze dynamic simulations of movement.
\newblock {\em IEEE transactions on biomedical engineering}, 54(11):1940--1950, 2007.

\bibitem{dwivedi2024poco}
Sai~Kumar Dwivedi, Cordelia Schmid, Hongwei Yi, Michael~J Black, and Dimitrios Tzionas.
\newblock Poco: 3d pose and shape estimation using confidence.
\newblock In {\em International conference on 3D vision (3DV)}, 2024.

\bibitem{fleisig2022comparison}
Glenn~S Fleisig, Jonathan~S Slowik, Derek Wassom, Yuki Yanagita, Jasper Bishop, and Alek Diffendaffer.
\newblock Comparison of marker-less and marker-based motion capture for baseball pitching kinematics.
\newblock {\em Sports Biomechanics}, pages 1--10, 2022.

\bibitem{ghorbani2021movi}
Saeed Ghorbani, Kimia Mahdaviani, Anne Thaler, Konrad Kording, Douglas~James Cook, Gunnar Blohm, and Nikolaus~F Troje.
\newblock Movi: A large multi-purpose human motion and video dataset.
\newblock {\em Plos one}, 16(6):e0253157, 2021.

\bibitem{gower1975generalized}
John~C Gower.
\newblock Generalized procrustes analysis.
\newblock {\em Psychometrika}, 40:33--51, 1975.

\bibitem{YOLOv8}
Glenn Jocher, Chaurasia Ayush, and Jing Qiu.
\newblock Yolo by ultralytics.

\bibitem{keller2023skin}
Marilyn Keller, Keenon Werling, Soyong Shin, Scott Delp, Sergi Pujades, C~Karen Liu, and Michael~J Black.
\newblock From skin to skeleton: Towards biomechanically accurate 3d digital humans.
\newblock {\em ACM Transactions on Graphics (TOG)}, 42(6):1--12, 2023.

\bibitem{keller2022osso}
Marilyn Keller, Silvia Zuffi, Michael~J Black, and Sergi Pujades.
\newblock Osso: Obtaining skeletal shape from outside.
\newblock In {\em Proceedings of the IEEE/CVF Conference on Computer Vision and Pattern Recognition}, pages 20492--20501, 2022.

\bibitem{kingma2014adam}
Diederik~P Kingma and Jimmy Ba.
\newblock Adam: A method for stochastic optimization.
\newblock {\em arXiv preprint arXiv:1412.6980}, 2014.

\bibitem{li2021lifting}
Wenhao Li, Hong Liu, Runwei Ding, Mengyuan Liu, and Pichao Wang.
\newblock Lifting transformer for 3d human pose estimation in video.
\newblock {\em arXiv preprint arXiv:2103.14304}, 2, 2021.

\bibitem{li2022exploiting}
Wenhao Li, Hong Liu, Runwei Ding, Mengyuan Liu, Pichao Wang, and Wenming Yang.
\newblock Exploiting temporal contexts with strided transformer for 3d human pose estimation.
\newblock {\em IEEE Transactions on Multimedia}, 2022.

\bibitem{li2022mhformer}
Wenhao Li, Hong Liu, Hao Tang, Pichao Wang, and Luc Van~Gool.
\newblock Mhformer: Multi-hypothesis transformer for 3d human pose estimation.
\newblock In {\em Proceedings of the IEEE/CVF Conference on Computer Vision and Pattern Recognition}, pages 13147--13156, 2022.

\bibitem{liu2020attention}
Ruixu Liu, Ju Shen, He Wang, Chen Chen, Sen-ching Cheung, and Vijayan Asari.
\newblock Attention mechanism exploits temporal contexts: Real-time 3d human pose reconstruction.
\newblock In {\em Proceedings of the IEEE/CVF Conference on Computer Vision and Pattern Recognition}, pages 5064--5073, 2020.

\bibitem{AMASS:ICCV:2019}
Naureen Mahmood, Nima Ghorbani, Nikolaus~F. Troje, Gerard Pons-Moll, and Michael~J. Black.
\newblock {AMASS}: Archive of motion capture as surface shapes.
\newblock In {\em International Conference on Computer Vision}, pages 5442--5451, Oct. 2019.

\bibitem{mundermann2006evolution}
Lars M{\"u}ndermann, Stefano Corazza, and Thomas~P Andriacchi.
\newblock The evolution of methods for the capture of human movement leading to markerless motion capture for biomechanical applications.
\newblock {\em Journal of neuroengineering and rehabilitation}, 3(1):1--11, 2006.

\bibitem{newell2016stacked}
Alejandro Newell, Kaiyu Yang, and Jia Deng.
\newblock Stacked hourglass networks for human pose estimation.
\newblock In {\em Computer Vision--ECCV 2016: 14th European Conference, Amsterdam, The Netherlands, October 11-14, 2016, Proceedings, Part VIII 14}, pages 483--499. Springer, 2016.

\bibitem{nguyen2023deep}
Kien~X Nguyen, Liying Zheng, Ashley~L Hawke, Robert~E Carey, Scott~P Breloff, Kang Li, and Xi Peng.
\newblock Deep learning-based estimation of whole-body kinematics from multi-view images.
\newblock {\em Computer Vision and Image Understanding}, page 103780, 2023.

\bibitem{Pagnon2022JOSS}
David Pagnon, Mathieu Domalain, and Lionel Reveret.
\newblock Pose2sim: An open-source python package for multiview markerless kinematics.
\newblock {\em Journal of Open Source Software}, 2022.

\bibitem{SMPL-X:2019}
Georgios Pavlakos, Vasileios Choutas, Nima Ghorbani, Timo Bolkart, Ahmed A.~A. Osman, Dimitrios Tzionas, and Michael~J. Black.
\newblock Expressive body capture: {3D} hands, face, and body from a single image.
\newblock In {\em Proceedings IEEE Conf. on Computer Vision and Pattern Recognition (CVPR)}, pages 10975--10985, 2019.

\bibitem{pavllo20193d}
Dario Pavllo, Christoph Feichtenhofer, David Grangier, and Michael Auli.
\newblock 3d human pose estimation in video with temporal convolutions and semi-supervised training.
\newblock In {\em Proceedings of the IEEE/CVF conference on computer vision and pattern recognition}, pages 7753--7762, 2019.

\bibitem{rajagopal2016full}
Apoorva Rajagopal, Christopher~L Dembia, Matthew~S DeMers, Denny~D Delp, Jennifer~L Hicks, and Scott~L Delp.
\newblock Full-body musculoskeletal model for muscle-driven simulation of human gait.
\newblock {\em IEEE transactions on biomedical engineering}, 63(10):2068--2079, 2016.

\bibitem{seth2018opensim}
Ajay Seth, Jennifer~L Hicks, Thomas~K Uchida, Ayman Habib, Christopher~L Dembia, James~J Dunne, Carmichael~F Ong, Matthew~S DeMers, Apoorva Rajagopal, Matthew Millard, et~al.
\newblock Opensim: Simulating musculoskeletal dynamics and neuromuscular control to study human and animal movement.
\newblock {\em PLoS computational biology}, 14(7):e1006223, 2018.

\bibitem{sigal2010humaneva}
Leonid Sigal, Alexandru~O Balan, and Michael~J Black.
\newblock Humaneva: Synchronized video and motion capture dataset and baseline algorithm for evaluation of articulated human motion.
\newblock {\em International journal of computer vision}, 87(1-2):4, 2010.

\bibitem{song2023markerless}
Ke Song, Todd~J Hullfish, Rodrigo~Scattone Silva, Karin~Gr{\"a}vare Silbernagel, and Josh~R Baxter.
\newblock Markerless motion capture estimates of lower extremity kinematics and kinetics are comparable to marker-based across 8 movements.
\newblock {\em Journal of Biomechanics}, 157:111751, 2023.

\bibitem{spurr2020weakly}
Adrian Spurr, Umar Iqbal, Pavlo Molchanov, Otmar Hilliges, and Jan Kautz.
\newblock Weakly supervised 3d hand pose estimation via biomechanical constraints.
\newblock In {\em Computer Vision--ECCV 2020: 16th European Conference, Glasgow, UK, August 23--28, 2020, Proceedings, Part XVII 16}, pages 211--228. Springer, 2020.

\bibitem{uhlrich2022opencap}
Scott~D Uhlrich, Antoine Falisse, {\L}ukasz Kidzi{\'n}ski, Julie Muccini, Michael Ko, Akshay~S Chaudhari, Jennifer~L Hicks, and Scott~L Delp.
\newblock Opencap: 3d human movement dynamics from smartphone videos.
\newblock {\em bioRxiv}, pages 2022--07, 2022.

\bibitem{van2018accuracy}
Eline Van~der Kruk and Marco~M Reijne.
\newblock Accuracy of human motion capture systems for sport applications; state-of-the-art review.
\newblock {\em European journal of sport science}, 18(6):806--819, 2018.

\bibitem{wu2021graph}
Size Wu, Sheng Jin, Wentao Liu, Lei Bai, Chen Qian, Dong Liu, and Wanli Ouyang.
\newblock Graph-based 3d multi-person pose estimation using multi-view images.
\newblock In {\em Proceedings of the IEEE/CVF international conference on computer vision}, pages 11148--11157, 2021.

\bibitem{xu2020deep}
Jingwei Xu, Zhenbo Yu, Bingbing Ni, Jiancheng Yang, Xiaokang Yang, and Wenjun Zhang.
\newblock Deep kinematics analysis for monocular 3d human pose estimation.
\newblock In {\em Proceedings of the IEEE/CVF Conference on computer vision and Pattern recognition}, pages 899--908, 2020.

\bibitem{929617}
B.-M. You, P. Siy, W. Anderst, and S. Tashman.
\newblock In vivo measurement of 3-d skeletal kinematics from sequences of biplane radiographs: Application to knee kinematics.
\newblock {\em IEEE Transactions on Medical Imaging}, 20(6):514--525, 2001.

\bibitem{zamir2022restormer}
Syed~Waqas Zamir, Aditya Arora, Salman Khan, Munawar Hayat, Fahad~Shahbaz Khan, and Ming-Hsuan Yang.
\newblock Restormer: Efficient transformer for high-resolution image restoration.
\newblock In {\em Proceedings of the IEEE/CVF Conference on Computer Vision and Pattern Recognition}, pages 5728--5739, 2022.

\bibitem{zhang2021direct}
Jianfeng Zhang, Yujun Cai, Shuicheng Yan, Jiashi Feng, et~al.
\newblock Direct multi-view multi-person 3d pose estimation.
\newblock {\em Advances in Neural Information Processing Systems}, 34:13153--13164, 2021.

\bibitem{zhang2022mixste}
Jinlu Zhang, Zhigang Tu, Jianyu Yang, Yujin Chen, and Junsong Yuan.
\newblock Mixste: Seq2seq mixed spatio-temporal encoder for 3d human pose estimation in video.
\newblock In {\em Proceedings of the IEEE/CVF Conference on Computer Vision and Pattern Recognition}, pages 13232--13242, 2022.

\bibitem{zheng20213d}
Ce Zheng, Sijie Zhu, Matias Mendieta, Taojiannan Yang, Chen Chen, and Zhengming Ding.
\newblock 3d human pose estimation with spatial and temporal transformers.
\newblock In {\em Proceedings of the IEEE/CVF International Conference on Computer Vision}, pages 11656--11665, 2021.

\end{thebibliography}
}


\end{document}